\documentclass[manuscript]{acmart}
\makeatletter

\renewcommand{\ACM@sect@format}[1]{#1}

\makeatother
\usepackage{CJKutf8}

\AtBeginDocument{%
  }

\setcopyright{acmlicensed}
\copyrightyear{2018}
\acmYear{2018}
\acmDOI{XXXXXXX.XXXXXXX}

\acmConference[Conference acronym 'XX]{Make sure to enter the correct
  conference title from your rights confirmation emai}{June 03--05,
  2018}{NY}
\acmISBN{978-1-4503-XXXX-X/18/06}

\begin{document}
% \begin{CJK*}{UTF8}{gbsn}
\title{Beyond Explicit Intent: Discovering and Shaping Artistic Intent through Human-AI Co-Creation}

\author{Renye Yan}
\authornote{Renye Yan, Jialun Zhong, and Jikang Cheng contributed equally to this work.}
\affiliation{%
  \institution{Peking University}
  \city{Beijing}
  \country{China}
}
\author{Jialun Zhong}
\authornotemark[1]
\affiliation{%
  \institution{Peking University}
  \city{Beijing}
  \country{China}
}
\author{Jikang Cheng}
\authornotemark[1]
\affiliation{%
  \institution{Peking University}
  \city{Beijing}
  \country{China}
}
\author{Yi Gao}
\affiliation{%
  \institution{Peking University}
  \city{Beijing}
  \country{China}
}
\author{Yanzeng Li}
\affiliation{%
  \institution{Peking University}
  \city{Beijing}
  \country{China}
}
\author{Ling Liang}
\authornote{Ling Liang is the corresponding author.}
\affiliation{%
  \institution{Peking University}
  \city{Beijing}
  \country{China}
}

% Short author list used in page headers.

% \renewcommand{\shortauthors}{Yan et al.}

\begin{abstract}
Generative AI is becoming an important medium for artistic creation, yet existing interactions often assume that creators can specify intentions in advance and accurately communicate them through text prompts. Artistic intentions, however, may be vague, latent, or recognizable only after seeing concrete visual possibilities. We propose a human-AI co-creation paradigm beyond explicit intent expression. By expanding model exploration, we expose artists to diverse visual possibilities, allowing them to identify previously unarticulated preferences through observation, comparison, and selection. We then integrate artists' curatorial judgments into reinforcement learning, assigning different learning strengths based on preferences over generation trajectories, so recognized directions are more strongly reinforced and the model gradually adapts to individual aesthetics. This creates a continuous ``explore-identify-shape-re-explore" loop in which artistic intent is progressively discovered, externalized, and shaped. Through three user studies, we examine latent-preference discovery, personalized preference alignment, and the externalization of vague, difficult-to-express creative intentions.

\end{abstract}




% 
\begin{teaserfigure}
  \includegraphics[width=\textwidth]{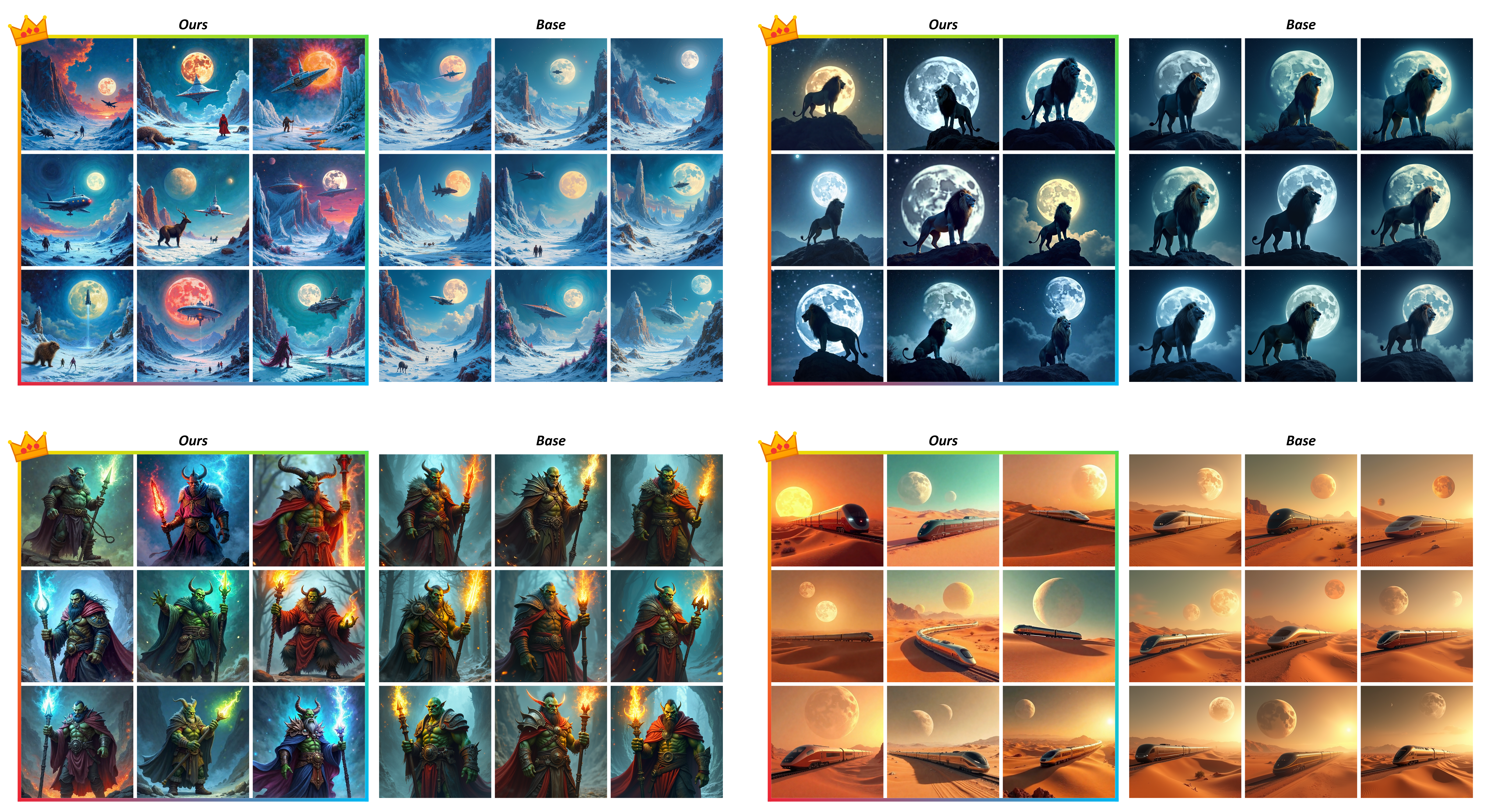}
  \caption{\textbf{Motivation}. Artists’ creative intentions do not always fully exist before generation. Through broader exploration, Ours externalizes potential tendencies that originally exist only as vague feelings or have not yet entered explicit awareness into concrete candidates, enabling creators to gradually realize what they may truly want after seeing the results.}
  % \Description{Enjoying the baseball game from the third-base
  % seats. Ichiro Suzuki preparing to bat.}
  \label{fig:teaser}
\end{teaserfigure}


\maketitle

\section{Introduction}
\label{sec:intro}

Generative AI~\cite{huang2024plantography,chen2024chatscratch,xiao2024typedance,choi2024creativeconnect,son2024genquery,dhawka2024better,rajcic2024towards,zhang2024partiality,lee2024impact,kamali2025characterizing,porquet2025copying,smith2026noise,zhao2026less,grace2026entropic,li2026preference,simonen2026exploration,wang2023reprompt,moore2023failurenotes}is rapidly entering visual art and design practice, gradually evolving from a tool for one-shot content generation into an important medium through which creators ideate, explore, and iterate. Text-to-image models can generate a large number of visual candidates within a short period of time, enabling creators to rapidly experiment with different compositions, styles, and expressive directions. At the same time, this new mode of creation is also changing the interaction between people and generative systems. Creators no longer simply execute a predefined production process; instead, they work together with models through an iterative cycle of proposing ideas, observing generated results, reconsidering their judgments, and adjusting their creative directions~\cite{huang2024plantography,cavallin2026designers}. Recent HCI research has begun to examine this shift, including how to help users improve and refine prompts, how to continuously incorporate user intent during model fine-tuning, and how to support designers in exploring, iterating on, and tracking multiple generative design alternatives~\cite{zeng2024intenttuner,peng2026designtrace}.

However, these works often still rely on an important underlying assumption: that users’ creative intentions already exist, and that the role of an interactive system is to help users express, communicate, or realize those intentions more accurately. For example, prompt-centered systems primarily address how users can translate existing ideas into language that models can understand, while intent-aware fine-tuning further helps users incorporate existing visual or textual intentions into model training~\cite{zeng2024intenttuner,mahdavi2024ai}. Even in generative design tools that support exploration across multiple alternatives, exploration is typically treated as a means for helping users compare and refine existing design directions~\cite{peng2026designtrace}. Such research has substantially improved communication between people and generative models, but it leaves open a more fundamental question: what if artists themselves cannot fully articulate what they want at the beginning of the creative process?

This question is particularly important in artistic creation. Creative intent is not always a fixed objective that is fully formed before generation and simply waits to be encoded into a prompt. Artists may begin with only a vague direction, emotion, or aesthetic inclination. They may also be unable to describe a particular visual form in words, yet immediately recognize, once they see it, that ``this is close to what I want'' or ``this is not the direction I am looking for.'' In other words, artistic creation involves not only the execution of existing intentions, but also the discovery, interpretation, and development of intentions that have not yet fully formed. Recent studies of generative AI in design practice have observed that creators advance their creative processes by repeatedly viewing and reflecting on AI-generated outputs~\cite{cavallin2026designers,naqvi2025catalyst}. Research on long-term artistic practice further suggests that creative and aesthetic relationships between artists and machine learning systems may gradually emerge through sustained interaction, rather than being fully predetermined by either side~\cite{naqvi2025catalyst}.

Building on this observation, we pose a research question that differs from asking how creative intent can be expressed more accurately: can generative AI help creators discover artistic intentions that have not yet been explicitly expressed, or even fully recognized, by actively expanding the visual possibilities available to them? We reinterpret exploration in generative models, moving beyond its conventional role as a model-side search mechanism and instead viewing it as a preference and intent discovery mechanism for creators. When a model generates highly similar results only around the current prompt, the range within which an artist can make judgments is also constrained. In contrast, when the system deliberately expands the generation space and presents more differentiated candidates, artists are exposed to visual directions that they may not have previously considered~\cite{naqvi2025catalyst,rozental2026neither}. These outputs do not necessarily need to become final artworks. Rather, they constitute an externalized space of possibilities through which artists can gradually recognize their latent aesthetic preferences by seeing, comparing, and selecting among alternatives~\cite{li2026preference}.

Making latent preferences visible, however, is not sufficient to support sustained human–AI co-creation. If an artist’s judgment remains limited to the current generation round, the model may still return to the same generic generation distribution in the next round. We therefore further connect artists’ selections directly to the model learning process. Specifically, we adopt artist-in-the-loop reinforcement learning, in which artists’ aesthetic judgments over different rollouts become trajectory-level learning signals. For highly preferred results, the model learns from the full reward; for partially preferred results, the reward is discounted according to the strength of preference; and results that are not preferred are excluded from the current policy update. In this way, artists do not need to formalize a complete aesthetic objective in advance. Instead, through continued selection, they can gradually determine which generation behaviors are worth reinforcing. As these judgments accumulate, a general-purpose generative model progressively adapts toward the aesthetic space of an individual artist.

This mechanism ultimately forms a continuous human–AI co-creation loop. Through exploration, the model expands the creative space that the artist can perceive; through observation and selection, the artist identifies directions that resonate with them; these selections then reshape the model’s generation distribution; and the updated model produces new visual possibilities for the artist to continue exploring. In our framework, artistic intent is therefore no longer treated as a fixed input that must exist before human–AI interaction begins. Instead, it can gradually emerge and evolve through a cycle of exploration–recognition–shaping–re-exploration. This perspective moves personalized generation beyond ``making the model more accurately execute my existing requirements'' toward ``enabling people and models to jointly discover what is worth further developing.''

To understand the role of this paradigm in human–AI co-creation, we conducted three user studies. Study 1 examines whether broader generative exploration can help artists reveal their latent artistic preferences. Study 2 further investigates whether artists’ continued selections can, through model learning, lead to stronger personalized preference alignment. Study 3 focuses on the broader creative implications of this process, examining whether sustained human–AI co-creation can help artists gradually externalize creative intentions that were previously latent, vague, and difficult to articulate, and turn them into visible visual expressions.

In summary, this paper makes the following contributions:

\begin{itemize}
\item We introduce a human–AI co-creation perspective beyond explicit intent, treating artistic intent as a dynamic object that can be discovered and shaped through sustained interaction. We further reinterpret generative exploration as a mechanism for helping creators reveal latent preferences and creative directions.

\item We develop an artist-in-the-loop reinforcement learning framework in which artists’ graded judgments over generation trajectories regulate model learning, allowing curatorial selection to continuously shape a personalized generative model.

\item Through three user studies, we investigate this interaction paradigm at three levels: preference discovery, personalized alignment, and the externalization of latent intent. Our findings provide empirical insights and design implications for supporting more exploratory, individualized, and sustained human–AI co-creation.

\end{itemize}

\section{Related Work}
This paper focuses on a question that has not yet been sufficiently studied: when creators’ own intentions remain vague or have not yet fully formed, can generative systems help such intentions gradually emerge through exploration and continuous learning, and further develop into generative systems adapted to individual artists? Building on this question, we review related work from three directions: Human–AI co-creation, personalization of generative models, and human feedback.

\subsection{Human--AI Co-Creation with Generative Models}
Generative models~\cite{cheng2026infodense,ma2026can,li2026videococo,ling2026gentrack} are increasingly becoming creative media in visual art and design practice. Unlike traditional tools that primarily execute explicit operations, generative systems can produce a large number of diverse visual candidates from limited input, allowing creators to continuously compare generated results and adjust their creative directions~\cite{feng2023promptmagician,peng2025fusain}. Early research mainly focused on lowering the barrier to text-to-image interaction, with one central question being how to help users translate existing creative ideas into prompts that models can effectively respond to. Subsequently, a growing body of HCI work has expanded its focus from one-shot prompt input to more sustained interaction processes~\cite{suh2024luminate,jain2025adaptivesliders}. For example, systems can help users revise prompts, support style exploration and local refinement, and allow creators to compare and revisit multiple design branches.

These studies have substantially enriched the interaction mechanisms available in generative creation, but their primary focus remains on how to better express, refine, and realize creative intentions that users have already formed~\cite{peng2025fusain,liu20233dall}. However, another situation in artistic creation has received comparatively less attention: creators may have only a vague sense of what they want, or may not even know that a particular direction is what they are looking for until it actually appears in visual form. In such cases, the creative process does not necessarily proceed from a clear intention toward its realization. Instead, the intention itself may gradually take shape through exposure to different visual possibilities.

Different from existing work on Human–AI co-creation, this paper introduces exploration and treats it as a creator-oriented intent discovery mechanism. By actively expanding the visible generation space, the system enables artists to encounter visual possibilities that go beyond the scope of their current explicit expressions. Through a process of “seeing–comparing–selecting,” artists can gradually recognize artistic preferences and creative directions that were previously difficult to articulate or had not yet been consciously recognized.

\subsection{Human Feedback for Interactive Model Adaptation}
Human feedback is an important mechanism for connecting user intent with the behavior of generative systems. Existing generative interaction systems commonly allow users to revise current outputs by modifying prompts, selecting candidate results, or adjusting local content. With the development of fine-tuning and preference optimization, such feedback has also begun to be incorporated into model training, allowing user judgments to further influence the model’s future generation behavior~\cite{lee2023aligning,xu2023imagereward,wallace2024diffusion}.

In existing approaches, human feedback is commonly transformed into pairwise comparisons, rankings, or reward signals that indicate which outputs better match expected outcomes~\cite{kirstain2023pick}. However, such feedback is still largely treated as correction or annotation and is used to serve predefined optimization objectives.

This paper further treats human feedback as a continuous model-shaping mechanism. Artists’ judgments over different rollouts directly determine their learning strength during reinforcement learning updates, thereby influencing the visual possibilities that the model can generate in subsequent rounds. In this way, feedback is used not only to evaluate current outputs, but also to continuously shape model behavior, forming a closed loop in which artist judgment and model adaptation interact with each other.

\subsection{Personalizing Generative Models to Human Preferences}
As generative models are increasingly used in personal creative practices, adapting general-purpose models to different users has become an important research direction. Early work on personalization primarily focused on enabling models to learn specific people, objects, or visual styles, so that generated outputs could consistently preserve a target concept. In recent years, this problem has gradually expanded toward preference personalization, where models need to understand not only what users want to generate, but also what different users prefer in generated results~\cite{zhang2025multi,puadurean2026inference,wang2026premier}.

This shift is particularly important because preference optimization in many generative models is built on population-level evaluations. Reward models or preference models~\cite{zhong2025comprehensive} trained on large-scale user data can improve visual quality in an overall sense, but population-level average preferences cannot fully represent the aesthetics of a particular creator. Different artists may have substantially different judgments about composition, color, and visual expression. Recent work has therefore begun to explore the use of individual preference samples to further adapt generative models, shifting from population-level alignment toward individualized preference alignment.

Some HCI research has pointed out that model personalization should not be understood solely as an offline parameter optimization problem. Users often need to participate in the model adaptation process and continuously express their goals during actual use. Some work therefore incorporates user intent into the fine-tuning workflow, allowing creators to guide data construction and model evaluation through textual or visual examples. These approaches enable models to account for user needs more directly, but they typically still require users to provide relatively explicit goals, examples, or preference data in advance.

For open-ended artistic creation, however, this assumption does not always hold. Artists may not be able to fully define their aesthetic space before training begins. Instead, they may gradually discover what truly aligns with their creative inclinations only after repeatedly encountering new generated results.

A key distinction between our work and existing personalized generation approaches is that we do not treat individual preference as a static target that must first be explicitly defined and then fitted by the model. Instead, artists’ preferences gradually emerge through interaction and selection, while these selections are continuously incorporated into model training, allowing preference discovery and model personalization to occur simultaneously. The resulting system is therefore not a one-time preference adaptation, but a generative system that progressively adapts to an individual artist’s aesthetics and creative inclinations through their continued judgments.

% Local fix for acmart authordraft + CJKutf8: force draft watermark text back to Latin font
\ifdefined\SetWatermarkText
\SetWatermarkText{{\normalfont\rmfamily\parbox{12em}{\centering
Unpublished working draft.\\
Not for distribution.}}}
\fi

% ============================================================
% Local fix: make acmart \subsubsection a block heading.
% This prevents CJKutf8 text after a subsubsection from inheriting
% acmart's run-in sans-serif/italic heading font and falling back
% to the missing Cyberbit font (e.g. cyberb4e).
% This definition only affects this Method file and later content
% while this definition remains active.
% ============================================================
\makeatletter
\renewcommand\subsubsection{%
  \def\@toclevel{3}%
  \@startsection{subsubsection}{3}{\z@}%
  {-.5\baselineskip \@plus -2\p@ \@minus -.2\p@}%
  {.25\baselineskip}%
  {\ACM@NRadjust\@subsubsecfont}%
}
\makeatother

\section{From Exploration to Artistic Intent Discovery and Model Shaping}

Intent in open-ended artistic creation is not always fully defined before interaction begins. Artists may only perceive a vague direction, or may gradually realize which forms truly align with their aesthetics only after seeing concrete visual results. Based on this observation, we design the system as a continuous exploration–recognition–shaping loop. The model first presents differentiated creative possibilities through broader exploration, allowing artists to encounter visual directions beyond the scope of their current explicit expressions. Artists then identify preferences that resonate with them by observing and curating these results. These judgments are further transformed into learning signals that continuously shape the model’s generation behavior. As the model adapts, its subsequent generations create new spaces for exploration, allowing the artist’s intent discovery and the model’s personalized development to continuously progress within the same interaction process. Through this design, we do not treat artists’ preferences as fixed targets that must be fully specified in advance, but as dynamic signals that can gradually emerge through sustained human–AI co-creation and further shape the generative system.

\subsection{Expanding the Space of Creative Possibilities}
\label{sec:creative_exploration}

This study adopts a Flow Matching diffusion model~\cite{yan2025entropy,yan2026pixel,yan2026explore,yanlitexplorer,yanmultitune} as the generative backbone and introduces an intrinsic-motivation-based exploratory generation mechanism to increase behavioral coverage in the generation space. We further transform this model-side exploration process into creative possibilities that artists can observe and compare, thereby providing an interactive foundation for discovering latent preferences.

\subsubsection{Flow-based Reinforcement Learning Formulation}
\label{sec:rl_formulation}
Let $\mathbf{x}_0 \sim X_0$ denote a real image sample,
$\mathbf{x}_1 \sim X_1$ denote noise sampled from a Gaussian distribution.
Rectified Flow defines the intermediate state over $t\in[0,1]$ as
$
    \mathbf{x}_t
    =
    (1-t)\mathbf{x}_0
    +
    t\mathbf{x}_1.
$
The generative model learns a parameterized velocity field
$\mathbf{v}_{\theta}(\mathbf{x}_t,t,\mathbf{c})$,
where $\mathbf{c}$ denotes the creative condition currently provided by the artist.
The corresponding Flow Matching training objective is:
\begin{equation}
        \mathcal{L}(\theta)
    =
    \mathbb{E}_{t,\mathbf{x}_0\sim X_0,\mathbf{x}_1\sim X_1}
    \left[
        \left\|
        \mathbf{v}
        -
        \mathbf{v}_{\theta}
        (\mathbf{x}_t,t,\mathbf{c})
        \right\|^2
    \right],
\end{equation}
where the target velocity field is:
$
    \mathbf{v}
    =
    \mathbf{x}_1-\mathbf{x}_0.
$
During inference, the model starts from noise
$\mathbf{x}_1$
and progressively generates the final image
$\mathbf{x}_0$
along the learned velocity field.
We further formulate the iterative generation process as a Markov decision process
$
    \mathcal{M}
    =
    (\mathcal{S},\mathcal{A},\rho_0,P,R).
$
In this MDP,
the Flow Matching model is treated as a parameterized policy
$\pi_{\theta}$.
Given the creative condition (prompt) that the artist can currently express explicitly,
$\mathbf{c}$.
The state in the generation process is defined as
$
    \mathbf{s}_t
    \triangleq
    (\mathbf{c},t,\mathbf{x}_t),
$
where
$\mathbf{x}_t$
denotes the current generation state,
and
$t$
denotes the current generation timestep.
The action produced by the model under state
$\mathbf{s}_t$
is defined as the generated sample at the next step
$
    \mathbf{a}_t
    \triangleq
    \mathbf{x}_{t-1}.
$
Therefore, the policy corresponding to the generative model can be represented as:
$
    \pi_{\theta}
    (\mathbf{a}_t\mid\mathbf{s}_t)
    \triangleq
    p_{\theta}
    (
        \mathbf{x}_{t-1}
        \mid
        \mathbf{x}_t,
        \mathbf{c}
    ).
$
Under this representation,
the model does not generate the complete image as a single action,
but progressively produces
$\mathbf{x}_{t-1}$
during the continuous generation process with
$\mathbf{x}_0$,
ultimately formed by a sequence of generation decisions.
Given the action
$\mathbf{a}_t=\mathbf{x}_{t-1}$
,
the state transition is deterministic, and its transition dynamics are written as
$
    P(
        \mathbf{s}_{t+1}
        \mid
        \mathbf{s}_t,
        \mathbf{a}_t
    )
    \triangleq
    (
        \delta_{\mathbf{c}},
        \delta_{t-1},
        \delta_{\mathbf{x}_{t-1}}
    ),
$
where
$\delta_y$
denotes a Dirac delta distribution centered at $y$. That is, the creative condition
$\mathbf{c}$
remains unchanged within a generation trajectory,
the generation timestep progresses from $t$ to $t-1$,
and the new generation state is determined by the action
$\mathbf{x}{t-1}$.
The initial state distribution is defined as
$
    \rho_0(\mathbf{s}_0)
    \triangleq
    \left(
        p(\mathbf{c}),
        \delta_T,
        \mathcal{N}(\mathbf{0},\mathbf{I})
    \right),
$
where
$p(\mathbf{c})$
denotes the distribution of creative conditions,
and the generation process starts from Gaussian noise
$
    \mathbf{x}_1
    \sim
    \mathcal{N}(\mathbf{0},\mathbf{I})
$
.
Because the overall visual result of a generation trajectory can only be evaluated after the final image has been formed, the reward is provided only at the final generation step:

\begin{equation}
    R(\mathbf{s}_t,\mathbf{a}_t)
    \triangleq
    \begin{cases}
        r(\mathbf{x}_0,\mathbf{c}),
        & t=0,\\[3pt]
        0,
        & \text{otherwise},
    \end{cases}
\end{equation}
where
$r(\mathbf{x}_0,\mathbf{c})$
denotes the base generation reward obtained by the final generated result
$\mathbf{x}_0$
under the current creative condition
$\mathbf{c}$
.

\subsubsection{Intrinsic Motivation for Diverse Generation}
To explicitly enhance exploration during the generation process, we further construct intrinsic diversity rewards based on the changes between adjacent generation states. For the $t$-th generation step, the intrinsic reward is defined as:
\begin{equation}
        R_{\mathrm{div}}
    =
    \left\|
        \mathbf{x}_{t}
        -
        \mathbf{x}_{t-1}
    \right\|_2,
\end{equation}
where
$\mathbf{x}_{t}$ and $\mathbf{x}_{t-1}$
denote the intermediate states at two adjacent generation steps, respectively.
This reward characterizes the degree of change produced by the model at the current generation step relative to the previous state:
larger state changes correspond to stronger exploration signals,
thereby encouraging the model to avoid prematurely concentrating on local generation modes.
We then use the hyperparameter $\lambda$
to balance the intrinsic exploration reward
with the external reward obtained by the final image (artist feedback),
resulting in the overall trajectory-level reward used for reinforcement learning:

\begin{equation}
    R_{\text{train}}
    =
    r(\mathbf{x}_0,\mathbf{c})
    +
    \lambda
            R_{\text{div}},
\end{equation}

where
$r(\mathbf{x}_0,\mathbf{c})$
measures the quality of the final generated result under the current creative condition,
while
$ R*{\text{div}}$
encourages the model to maintain sufficient state changes throughout the generation process.
$\lambda$ controls the trade-off between generation quality and exploration strength.
Based on this reward,
for the same creative condition
$\mathbf{c}$,
we sample $G$ sufficiently diverse generation trajectories.
$
    \left\{
        \tau^i
    \right\}_{i=1}^{G}
    \sim
    \pi_{\theta}
    (\cdot\mid\mathbf{c}),
$
And obtain the corresponding candidate results.
$
    \mathcal{X}_{\mathbf{c}}
    =
    \left\{
        \mathbf{x}_0^1,
        \mathbf{x}_0^2,
        \ldots,
        \mathbf{x}_0^G
    \right\}.
$
Therefore, exploration does not arise solely from repeated sampling, but is explicitly incorporated into the policy optimization objective through the intrinsic reward. While maintaining the quality of the final generated results, the model is encouraged to explore generation trajectories with more substantial state changes, thereby presenting artists with a broader range of visual possibilities. These candidates then serve as concrete objects for artists to compare and identify their preferences, and provide the basis for subsequent artist-guided model shaping.

\subsection{Recognizing Artistic Preferences through Curation}
\label{sec:artist_curation}

The previous section expands the creative possibility space available to artists through exploration, but the value of different generation directions for a particular artist still needs to be judged by the artist themselves. Therefore, we treat curation as a
preference recognition mechanism,
enabling artists to gradually identify aesthetic tendencies that have not yet been explicitly expressed by observing and comparing concrete visual results.
For the same creative condition $\mathbf{c}$,
the system presents the artist with a candidate set obtained from $G$ generation trajectories
$
    \mathcal{X}_{\mathbf{c}}
    =
    \left\{
        \mathbf{x}^{1}_{0},
        \mathbf{x}^{2}_{0},
        \ldots,
        \mathbf{x}^{G}_{0}
    \right\}.
$

We adopt a
\textit{recognition-before-articulation}
interaction approach:
artists do not need to define a complete preference profile in advance,
nor do they need to explain why a particular result aligns with their aesthetics.
Instead, they only need to judge the extent to which each candidate is worth further developing.
For candidate $\mathbf{x}_{0}^{i}$,
its curatorial judgment is represented as
$
    y_i
    \in
    \{
        \texttt{Preferred},
        \texttt{Partially Preferred},
        \texttt{Not Preferred}
    \}.
$
The three-level feedback preserves uncertainty in artistic judgment.
Here,
\texttt{Preferred} indicates that the artist clearly approves of the current generation direction,
\texttt{Partially Preferred} indicates that the result does not yet fully meet expectations but contains creative potential worth retaining,
and \texttt{Not Preferred} indicates that the current direction does not align with the artist’s personal preferences.

Importantly, these judgments are not evaluations of the general quality of an image, but rather reflect artist-specific preference. The same generated result may receive different judgments from different artists, and such differences are precisely the information that personalized creation needs to preserve.

\subsection{Turning Artist Judgments into Learning Signals}
\label{sec:judgment_to_learning}

In the previous section, the artist provides graded judgments on different generation directions through curation:
\texttt{Preferred}, \texttt{Partially Preferred} and
\texttt{Not Preferred}.
To allow these judgments to further shape the model, we transform them into
rollout-level learning weights, so that the degree of the artist's preference directly determines
the learning strength of the corresponding generation trajectory in the current policy update.
This design follows
\textit{stronger preference leads to stronger reinforcement}.
Specifically,
for the $i$-th generation trajectory, let the artist's curatorial judgment be $y_i$,
we define its preference weight as:
\begin{equation}
w_i =
\begin{cases}
1, & y_i=\texttt{Preferred},\\[3pt]
\alpha, & y_i=\texttt{Partially Preferred},\\[3pt]
0, & y_i=\texttt{Not Preferred},
\end{cases}
\qquad 0<\alpha<1.
\end{equation}
where $\alpha$ controls the learning strength of partially preferred trajectories.
Combined with the
trajectory-level training reward $R_{\text{train}}$, 
the reward modulated by artist preference is:
\begin{equation}
    \widetilde{R}_{i}
    =
    w_i R_{\text{train}}.
\end{equation}
When $w_i=1$ , the reward signal of the trajectory is fully retained for policy optimization;
when $w_i=\alpha$, its learning effect is correspondingly weakened;
when $w_i=0$ , the trajectory does not participate in the current policy update.
Therefore, the artist's curation is no longer merely used to filter the current generated results,
but directly determines which generation behaviors should continue to be reinforced by the model.

Through the preceding process, the model first presents differentiated visual possibilities to the artist through exploration,
the artist then identifies directions that align with their aesthetics through curation,
and further transforms these judgments into preference-weighted reward
$\widetilde{R}_i$.
We use these artist-conditioned learning signals to update the generative model,
allowing the artist's local selections to gradually accumulate into long-term changes in the model's generation behavior.
The updated model does not end the current interaction, but is instead used again for the next round of generation.
Therefore, each round of curation changes the visual directions that the model is more likely to generate subsequently,
while the new generated results continue to provide new objects for the artist to evaluate.
As this process is repeatedly carried out, the originally general-purpose generative model gradually develops into a
personalized generative model that better aligns with the aesthetics and creative inclinations of a specific artist.
This adaptation does not mean that the model simply copies results that the artist has already selected.
The preceding intrinsic exploration mechanism is continuously retained in subsequent generation,
allowing the model to continue presenting new visual possibilities within the gradually personalized generation space.
Therefore, personalization accumulates the artist's historical judgments,
while exploration continuously preserves discovery space for creative directions that have not yet formed.

We plan to recruit 24 participants with sustained experience in visual creation for the subsequent studies. Participants will be between 18 and 45 years old and will be required to have at least 2 years of sustained visual creation experience. Their creative backgrounds may include digital art, visual design, illustration, and other creative practices centered on visual expression. Among them, 12 participants will have experience using text-to-image generative systems, and at least 8 will have relatively frequent experience creating with generative AI. Participants will be required to independently make visual judgments and creative decisions based on their own creative practice.

This recruitment strategy is aligned with the research goals of this paper. We are not concerned with general evaluations by ordinary users of which image looks better, but rather with whether creators can identify aesthetically meaningful tendencies with personal significance through sustained exposure to generated results, and further shape the model through these judgments. Therefore, we treat participants as creators with their own creative practices and aesthetic orientations, rather than as annotators of general image quality.

The three studies respectively focus on preference discovery, personalized preference alignment, and long-term creative intent development. Before entering the specific studies, participants will receive the same introduction to system operation, while procedures related to the specific tasks of each Study will be further described in the corresponding sections.

\section{Study 1: Does Exploration Help Reveal Artistic Preferences?}
Our first study aims to test the core hypothesis of this paper: whether broader generative exploration can help artists identify creative preferences that they have not previously expressed explicitly. Participants will experience both conventional generation and enhanced exploration conditions, and will compare preference discovery under the two conditions through behavioral choices, subjective evaluations, and interviews. Our focus is on whether these newly introduced visual possibilities genuinely help artists discover aesthetic tendencies that were previously latent or difficult to articulate.

\subsection{Study1 Design and Conditions}

We adopt a within-subject design, in which each participant experiences both Conventional Generation and Exploratory Generation. The two conditions use the same base model and keep generation quality, resolution, text conditions, interaction interface, and number of candidates consistent. No model updates are performed during the experiment, thereby limiting the primary experimental variable to the degree of exploration. Conventional Generation uses standard sampling, keeping candidates concentrated around the current prompt; Exploratory Generation uses the stochastic exploration mechanism proposed in Section 3 to expand the coverage of generation trajectories, allowing participants to encounter more differentiated visual possibilities. Each participant completes one open-ended visual creation task of comparable complexity under each condition, with both task assignment and condition order counterbalanced.

\subsection{Task and Procedure}

Before the experiment begins, we introduce participants to the system operation without explaining the research hypotheses. In each task, participants first enter an initial prompt and briefly record the creative intent that they can explicitly articulate before generation. They then browse the candidate results under the corresponding condition and select directions that are worth further developing. For preferences triggered by the candidate results but not yet explicitly recognized before generation, participants provide a brief description and restate their current creative intent after the exploration ends. By comparing pre-generation and post-exploration intent articulation, we examine whether exploration leads to the emergence of new preferences or makes previously vague intentions clearer. After each condition, participants complete a subjective questionnaire and a semi-structured interview, and compare the two experiences after completing both conditions.

\subsection{Measures and Analysis}

We focus on whether diversity genuinely changes artists’ preference discovery. The primary measures include participants’ subjectively perceived Preference Discovery; Intent Articulation Gain, derived from changes in intent articulation before and after generation; and Preference Recognition Rate, which captures the proportion of candidates that actually trigger the recognition of new preferences. Preference Discovery is assessed using a subjective scale, while Intent Articulation Gain is obtained by having two independent coders compare participants’ descriptions before and after generation.
We further measure Creative Possibility Coverage and Perceived Inspiration, while also recording Perceived Control and Cognitive Burden to evaluate the benefits and costs of exploration diversity.

\subsection{Study1 Results}
\begin{figure*}
    \centering
    \includegraphics[width=1\linewidth]{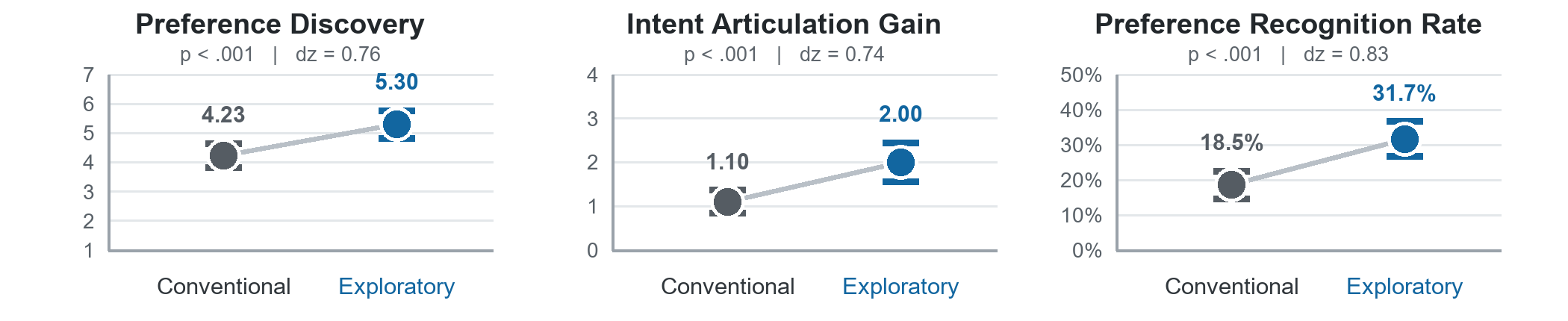}
    \vspace{-0.6cm}
    \caption{Comparison of Conventional and Exploratory Generation across the three primary outcomes in Study 1. Points indicate condition means and error bars denote 95\% confidence intervals. Exploratory Generation produced consistently stronger preference discovery outcomes across all measures.}
    \label{fig:study1_preference_discovery}
\end{figure*}

As shown in Figure~\ref{fig:study1_preference_discovery}, Exploratory Generation consistently produced stronger outcomes than Conventional Generation across Preference Discovery, Intent Articulation Gain, and Preference Recognition Rate. Participants were more likely to identify previously unarticulated preferences and to express newly clarified creative intentions after broader exploration. All three comparisons showed statistically significant improvements, supporting the role of exploration in facilitating preference discovery.

\subsection{Study1 Discussion}
Study 1 focuses on whether broader visual possibilities can facilitate artistic preference discovery. Its core significance lies in showing that exploration is not only a model-side search mechanism, but can also serve as a creator-oriented preference recognition mechanism: by exposing creators to more comparable visual directions, the system allows some preferences that were previously difficult to articulate or had not yet been consciously recognized to take on identifiable forms.

This result further challenges the common assumption that ``creators first clarify their intentions, and systems then implement them.'' In open-ended artistic creation, preferences may gradually emerge through a process of ``seeing--comparing--recognizing,'' rather than always existing prior to interaction. Therefore, the value of exploration is not to determine what artists should like, but to expand the creative possibility space within which they can form judgments.

\section{Study 2: Can Artist Curation Improve Personalized Preference Alignment?}

Study 2 further examines whether preferences identified by artists during exploration can be transformed into stable personalized model adaptation through continuous curation. Participants shape the model through multiple rounds of selection, after which we compare the Self-Personalized Model, Base Model, and Other-Artist Personalized Model in a blind evaluation to examine whether the model not only better aligns with the individual artist’s aesthetics, but also develops genuine artist-specific preference alignment.

\subsection{Study2 Design and Conditions}

To examine whether personalization genuinely reflects the aesthetics of an individual artist rather than merely bringing about a general improvement in quality, Study 2 adopts a within-subject design with three conditions: Base Model, Self-Personalized Model, and Other-Artist Personalized Model. The Self-Personalized Model is shaped by the participant’s own continuous curation, whereas the Other-Artist Personalized Model is shaped by feedback from another artist, with the same training budget controlled across conditions.

We use held-out prompts for blind evaluation. All models use identical generation settings, and candidates are presented anonymously and in randomized order, so that the comparison focuses on artist-specific preference alignment.

\subsection{Task and Procedure}

Study 2 includes a personalization phase and an evaluation phase. In the personalization phase, all participants start from the same base model. In each round, the system generates a set of exploratory candidates, and participants label them as Preferred, Partially Preferred, or Not Preferred according to their personal aesthetics. The system then updates the model according to the preference weighting described above. This process continues across multiple rounds of interaction, allowing participants’ sustained curation to gradually form the Self-Personalized Model.

After personalization is completed, we freeze the model and enter the blind evaluation phase. Based on held-out prompts, participants conduct pairwise comparisons between results generated by different models and judge which result better aligns with the creative direction they would like to continue developing. After completing all comparisons, participants fill out a subjective questionnaire and participate in a semi-structured interview to further understand their perceptions of personal fit and the model adaptation process.

\subsection{Measures and Analysis}
\label{sec:study2_measures}

Study 2 evaluates Personalized Preference Alignment mainly at two levels: behavioral choice and subjective perception. At the behavioral level, we record the Self-Model Preference Rate, separately comparing participants’ preference proportions for the Self-Personalized Model relative to the Base Model and the Other-Artist Personalized Model in the blind evaluation. The former is used to determine whether personalization improves individual preference matching, while the latter further examines whether this improvement is artist-specific rather than merely reflecting a general improvement in model quality. We also compare differences across artists in their preferences for their own models versus other artists’ models to examine whether the personalization effect exhibits stable individual characteristics.

Subjective evaluations include Perceived Personal Fit, Perceived Model Understanding, and Creative Control, which respectively measure the degree of alignment between the model and the participant’s personal aesthetics, participants’ perception that the model gradually ``understands'' their preferences, and their perceived control over the generation direction. We also record Perceived Variety and Novelty Preservation to examine whether personalization preserves sufficient generative diversity while improving personal fit.

For the pairwise preference data, we use mixed-effects logistic regression and control for repeated-measurement differences introduced by participant and prompt. For subjective scales, we use appropriate repeated-measures statistical tests according to the data distribution and report effect size and confidence interval. Semi-structured interviews are analyzed using thematic analysis to further understand how participants perceive changes in model personalization, as well as the potential benefits and limitations of personalization.

\subsection{Study2 Results}

\begin{figure*}
    \centering

    \begin{minipage}[t]{0.6\linewidth}
        \centering
        \includegraphics[width=\linewidth]{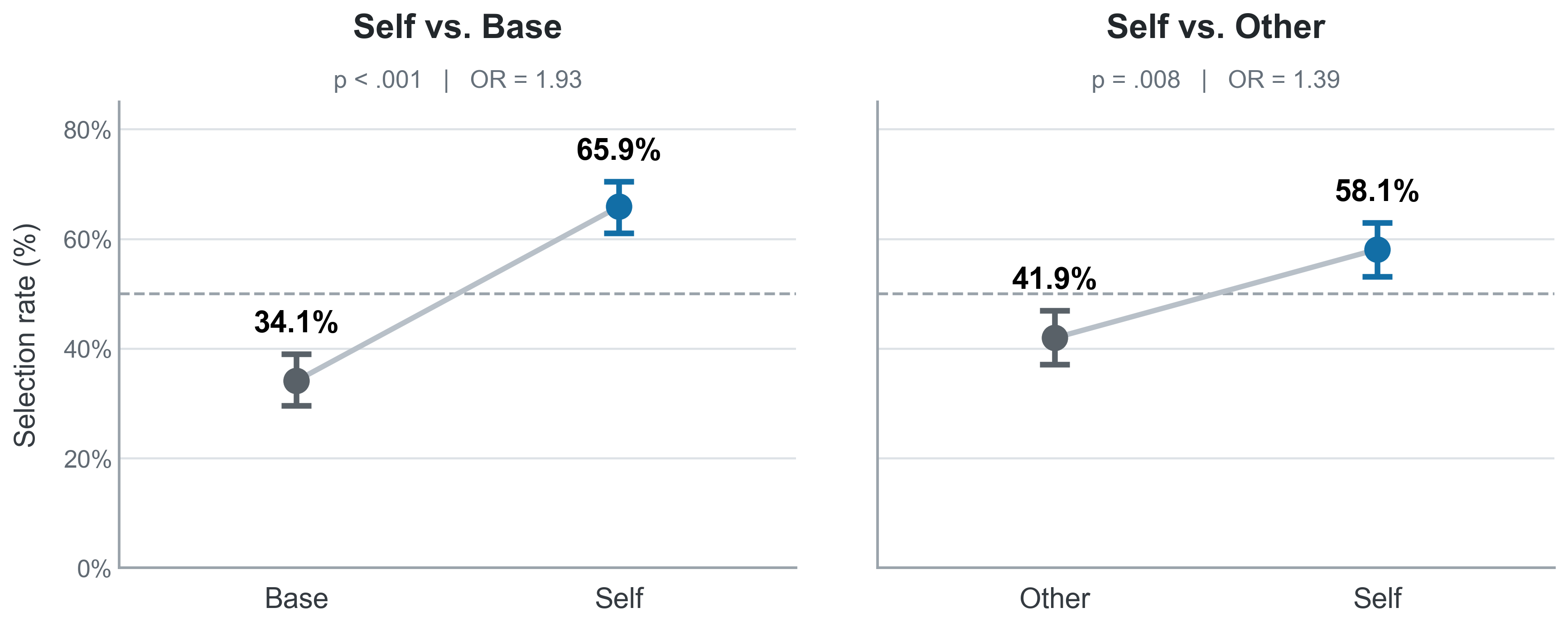}
        \caption{Blind pairwise evaluation of model personalization in Study 2. Points indicate the proportion of comparisons favoring the Self-Personalized Model, with error bars showing 95\% confidence intervals and the dashed line indicating chance level.}
        \label{fig:study2_preference_simple}
    \end{minipage}
    \hfill
    \begin{minipage}[t]{0.35\linewidth}
        \centering
        \includegraphics[width=\linewidth]{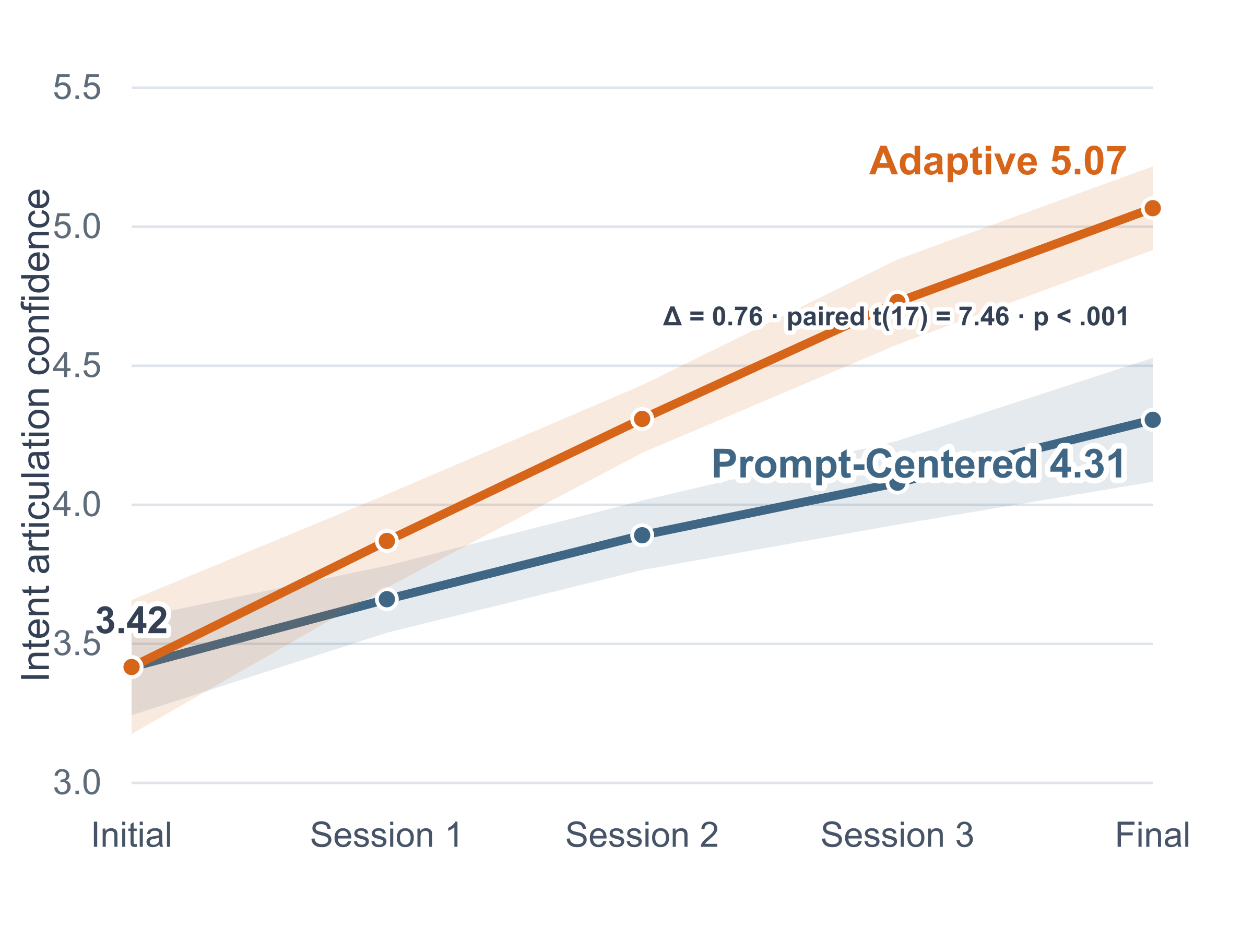}
        \caption{Changes in Intent Articulation Confidence across Study 3 sessions. Lines indicate condition means and shaded bands denote 95\% confidence intervals.}
        \label{fig:study3_publication_figure}
    \end{minipage}

\end{figure*}

As shown in Figure~\ref{fig:study2_preference_simple}, participants more frequently preferred their Self-Personalized Model over both the Base Model and the Other-Artist Personalized Model during blind evaluation. The advantage over the Base Model indicates improved personal preference alignment, while the preference over models personalized by other artists suggests that the learned adaptation retained an artist-specific component.

\subsection{Study2 Discussion}

Study 2 focuses on whether artists’ sustained curation can lead to genuine artist-specific personalization. If artists not only prefer their own personalized model over the Base Model, but also prefer it over models shaped by other artists, this would indicate that the system has learned individualized aesthetic judgments rather than producing a uniform improvement in quality.

This result further transforms curation from result selection into a model-shaping practice. Artists do not need to define a complete preference profile in advance. Instead, by continuously judging which directions are worth further developing, they can gradually shape the model’s future generation behavior. In other words, personalization can gradually emerge through curation rather than relying on prior specification.

\section{Study 3: Can human–AI co-creation externalize and materialize artists’ latent, ambiguous, and difficult-to-articulate creative intentions?}
Study 1 examines whether exploration helps artists discover latent preferences, while Study 2 investigates whether these preferences can continuously shape a personalized model through curation. Study 3 further situates both processes within sustained human–AI co-creation, examining whether creative intentions that were originally latent, vague, or difficult to articulate can gradually be externalized and take on concrete visual forms as artists interact over time with a generative system that continuously adapts to their judgments.

We define externalization as the process through which artists gradually recognize and express creative tendencies that were previously difficult to articulate, and materialization as the process through which these intentions further acquire perceptible visual forms. To study this process, we adopt a longitudinal mixed-method design in which participants engage in multi-session creation around open-ended themes that are personally meaningful but not yet fully determined. We analyze how artistic intent is discovered, revised, and concretized through sustained interaction by combining generated results, curation behaviors, changes in intent descriptions, and reflective interviews.

\subsection{Study3 Design and Conditions}

Study 3 adopts a longitudinal within-subject design comparing Prompt-Centered Co-Creation and Adaptive Co-Creation. In the former, participants interact with a fixed base model by modifying prompts, regenerating results, and selecting candidates; in the latter, the system further incorporates the complete exploration--recognition--shaping loop, allowing artists’ curation to continuously influence model updates and gradually form artist-specific generative behavior.

Both conditions support normal prompt refinement and result selection, with the core difference being whether the model continuously adapts to the artist’s judgments. Each participant completes the two conditions around two personal creative themes with a similar degree of openness, with theme assignment and condition order counterbalanced. Our focus is not on final image quality, but on whether different co-creation mechanisms change the processes through which artistic intent is formed, externalized, and materialized.

\subsection{Task and Procedure}

Before each creative task begins, participants first record an Initial Intent Statement describing the creative direction they can currently express, as well as aspects that remain vague or difficult to articulate. They then enter the corresponding co-creation condition and engage in multiple rounds of creation. Under the Adaptive Co-Creation condition, participants also curate candidates as Preferred, Partially Preferred, or Not Preferred, and the model is continuously updated based on these judgments.

Reflection checkpoints are included in each session to record whether participants discover new creative directions and whether their understanding of their current intent changes. At the end of each session, participants update their Current Intent Statement; after completing the final creation, they submit a Final Intent Statement and participate in an artifact-based semi-structured interview. By combining intent descriptions from different stages, generated results, and interaction traces, we track how artistic intent is discovered, revised, and gradually materialized through sustained co-creation.

\subsection{Measures and Analysis}

Study 3 adopts a qualitative-dominant mixed-method analysis, focusing on how artistic intent is externalized and gradually materialized through sustained co-creation. We first compare the Initial, Current, and Final Intent Statements to analyze whether previously vague creative tendencies become more explicit, and distinguish between intent clarification and intent emergence. We then combine these statements with generated artifacts to examine whether these gradually formed intentions further acquire perceptible visual expressions, namely Intent Materialization.

To connect changes in intent with visual outcomes, we further trace intent--artifact trajectories by aligning intent descriptions, generated results, and interaction traces across different stages. At the same time, we record Intent Articulation Confidence, Perceived Agency, Creative Ownership, and Perceived Co-Adaptation to compare differences between the two co-creation conditions in intent expression, agency, and co-adaptation.

\subsection{Study3 Results}

As shown in Figure~\ref{fig:study3_publication_figure}, Intent Articulation Confidence increased across sessions under both conditions, but the increase was substantially stronger under Adaptive Co-Creation. The two conditions began at comparable levels and gradually diverged over time, with Adaptive Co-Creation producing significantly higher confidence by the final session. This pattern suggests that continued model adaptation can better support the progressive clarification and articulation of ambiguous creative intentions.

\subsection{Study3 Discussion}

Study 3 extends the problem addressed in this paper from preference recognition and model personalization toward longer-term creative intent development. Its core significance lies in recognizing that artistic intent is not always a fixed goal that is fully formed before interaction and simply waits to be executed by the system. Some intentions may only gradually become recognizable, expressible, and ultimately take on visual form after concrete visual possibilities emerge. Therefore, generated results are not only products of intent execution, but may also serve as materials for intent formation.

In this process, the role of personalization is not limited to improving preference alignment, but also to providing continuity for newly recognized creative directions. Artists gradually understand their own intentions through the possibilities proposed by the model, while the model changes the possibilities it can subsequently propose through sustained curation, thereby forming a bidirectional process of co-adaptation.

\begin{figure*}
    \centering
    \includegraphics[width=1\linewidth]{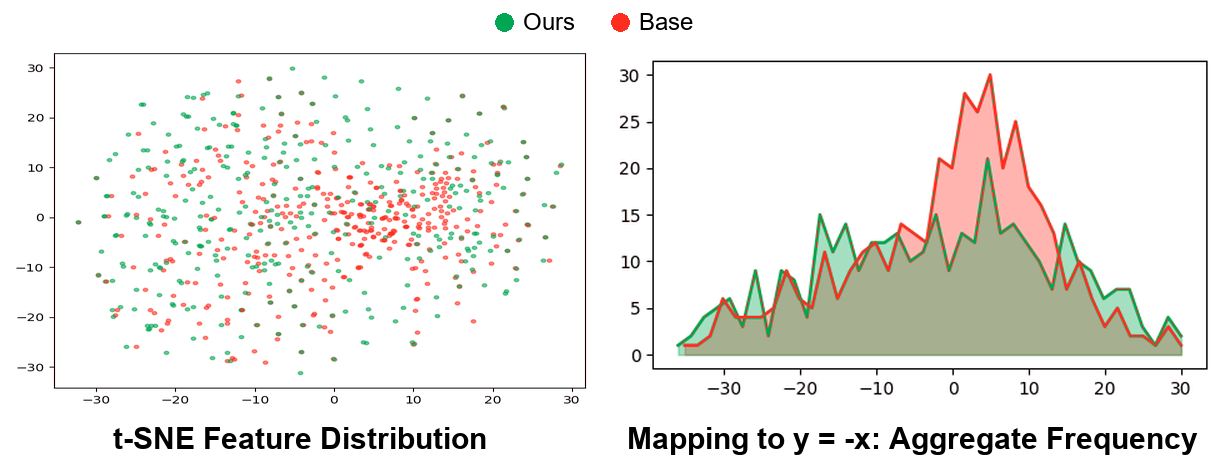}
\caption{Visualization of the generation-space distribution for Ours and Base. Left: t-SNE projection of generated features. Right: aggregate frequency after projecting the features onto $y=-x$. Compared with Base, Ours covers a broader region and exhibits a less concentrated distribution, visually indicating that our exploration mechanism expands the range of generative possibilities available to artists.}
\label{fig:distribution}
\end{figure*}

% Study 1

% Study 2 and Study 3

\section{Discussion}
Across the three studies, this paper is concerned not only with how generative systems understand and execute creative intentions that artists have already articulated, but also with whether they can help intentions that have not yet fully formed gradually acquire visible forms. In open-ended artistic creation, the creative directions in an artist’s mind may initially exist only as vague feelings, incomplete ideas, or even tendencies that have not yet entered explicit awareness. By continuously presenting concrete visual possibilities, generative models allow these previously difficult-to-describe or even difficult-to-notice creative tendencies to be seen, compared, and recognized by artists, and further developed into clearer creative directions through sustained interaction. In this sense, generative systems are no longer merely tools for realizing existing intentions, but can also become creative media that help latent intentions move from internal perceptions toward external visual forms.

\subsection{From Intent Expression to Intent Formation}
Existing generative interactions typically focus on helping users express and realize their existing intentions more accurately. However, our study further considers another situation: creators may not simply ``know what they want but be unable to articulate it.'' Instead, some creative directions may themselves not yet be fully formed and may only gradually become recognizable after concrete visual possibilities are presented.

Therefore, creative interaction should not be understood solely as intent transmission, but should also include intent formation. For creators whose goals remain unclear, the value of the system lies not only in narrowing the gap between prompts and expected outcomes, but also in providing sufficiently rich external materials through which unformed ideas can be observed, compared, and further developed. In this sense, ``Beyond Explicit Intent'' does not reject explicit expression, but rather recognizes that creative intent may simultaneously include parts that can already be articulated, parts that remain to be discovered, and parts that only gradually emerge through sustained creative practice.

\subsection{Exploration and Curation as Human-Facing Mechanisms}

This paper further reveals the human-facing value of exploration in generative creation. It not only serves model search, but also determines which visual possibilities artists are able to encounter. Only when a particular direction is actually presented can it become an object for creators to compare, reflect on, and form judgments about. Therefore, exploration expands not only the model’s search space, but also the creative possibility space within which artists can make aesthetic judgments.

Curation further transforms this recognition into model shaping. When artists’ selections can continuously influence model updates, curation no longer merely determines whether a current work is retained, but also determines which generative behaviors are worth further developing by the model. This mechanism allows personalization to gradually emerge from actual creative judgments, without requiring artists to first formalize complex aesthetic preferences into a complete preference profile. We therefore understand this process as personalization through curation rather than specification.

\subsection{Design Implications for Personalized Creative Systems}

The findings above suggest several important design directions for long-term personalized co-creation. First, systems should treat intent uncertainty as a normal state of creative practice rather than as an interaction failure that needs to be eliminated as quickly as possible. When users have not yet formed a clear direction, systems can support the discovery and formation of intent by presenting alternatives with meaningful differences.

Second, when users’ curation continuously changes the model, such long-term influence needs to remain understandable and controllable to creators. Users need to know which of their judgments are shaping the model and how their past selections influence current generation.

More broadly, long-term creative systems need to support not only preference learning, but also preference revision. Artists may want to reinforce certain historical directions, but they may also want to reduce the influence of past preferences in a new project and reopen the space for exploration. Therefore, the design problem for future creative AI will no longer be limited to helping users control ``what the model generates now,'' but will also include how to enable users to shape, inspect, revise, and selectively undo how the model learns from them.

From this perspective, a truly valuable personalized creative system is not one that can ultimately predict exactly what an artist will like, but one that can both sustain creative directions that have already been discovered and continuously preserve space for directions that have not yet formed.

\section {Limitations and Future Work}
Although this paper examines the roles of exploration, artist curation, and personalized model adaptation in human--AI co-creation through three interconnected user studies, our research still has several limitations. First, this paper primarily focuses on visual art and generative image creation, and whether the current findings can generalize to other creative practices requires further validation. Intent formation in artistic creation is clearly medium-dependent. Visual artists can quickly recognize whether a particular composition, atmosphere, or visual language resonates with them through concrete images, whereas intent in music creation, 3D design, or interactive storytelling may gradually emerge in different forms. Future work can extend the exploration--recognition--shaping loop proposed in this paper to other creative media and further investigate how different forms of generative feedback influence the process through which creators recognize and develop latent intentions.

Second, our system simplifies artists' curatorial judgments into three levels: Preferred, Partially Preferred, and Not Preferred. This design reduces interaction burden and allows artists' judgments to directly enter the reinforcement learning process, but it cannot fully express the more complex structure of artistic aesthetics. An artist may like the composition of an image but not its color; they may also wish to retain a particular visual direction while modifying only some local elements. Compressing such multidimensional judgments into trajectory-level preference may prevent the system from distinguishing which specific aspects of a generated result the artist actually approves of. Future systems can further support more fine-grained feedback, such as local region selection, attribute-level preferences, or natural language explanations, and investigate how these forms of feedback can be combined with trajectory-level learning without substantially increasing artists' interaction burden. One direction worth further exploring is to allow the system to actively ask for clarification when more explicit supervision is needed, rather than requiring artists to continuously provide costly annotations for all generated results.

Third, this paper interprets Not Preferred as ``not allowing the current trajectory to continue shaping the model,'' rather than as a universally negative aesthetic judgment. This design helps avoid incorrectly interpreting an individual artist's rejection as a general value judgment, but it also means that the model does not actively learn to move away from all rejected directions. Whether selective reinforcement alone is sufficient to form a stable personalized generation space over longer timescales remains an open question. Future work can compare different forms of preference learning, such as the current rollout filtering, negative preference optimization, or preferred--rejected pairwise learning, and examine how these mechanisms differ in personalization strength, generative diversity, and artists' sense of control.

Another important limitation comes from the persistent tension between personalization and exploration. The core design of this paper attempts to allow the model to gradually adapt to artists' selections while continuing to preserve its ability to propose new possibilities. However, as the model increasingly reinforces previously approved directions, the generation distribution may still gradually contract toward existing preferences. In conventional optimization, this phenomenon may be understood as convergence, but in open-ended artistic creation, overly rapid convergence may instead limit new intent discovery. In particular, when artists' aesthetics are themselves still changing, a model that relies too heavily on historical feedback may solidify temporary selections into a stable user profile. Future work should therefore focus not only on improving personalized preference alignment, but also on further investigating how to dynamically regulate exploration and exploitation. The system could adaptively expand or narrow the exploration range according to the artist's current intent certainty, creative stage, or recent feedback changes, thereby avoiding the personalization paradox of ``the more the system understands the user, the more it limits the user.''

This issue further involves user control in long-term human--AI co-creation. As generative models begin to retain and accumulate personal preferences, creators need not only to decide what the system should learn, but may also need to decide what the system should forget. The current model update mechanism in this paper mainly focuses on how to accumulate artists' curatorial judgments, but has not yet systematically investigated preference revision, temporary preference, or intentional forgetting. For example, an artist may consistently prefer a particular visual language in one project but may not want that preference to influence a completely different subsequent project. Future personalized creative systems therefore need to provide clearer preference boundaries and model memory control, enabling artists to inspect which historical selections are influencing current generation and allowing them to pause, reset, or reorganize those influences. Such control is not merely a system functionality issue, but is also directly related to creators' understanding of and agency over personalized models.

In addition, our user studies can still cover only a limited time window of the artistic creation process. Even though Study 3 adopts a sustained human--AI co-creation design, artists' aesthetic identities and creative habits typically change over much longer periods. Preference stabilization observed in short-term experiments does not necessarily indicate stable creative preferences in the long term, and may instead reflect participants' temporary adaptation to the current task. Future work should conduct longer-term in-the-wild deployments to examine whether artists still feel that the model ``understands them better'' after several weeks or even months of use, and whether such long-term adaptation produces dependence, aesthetic fixation, or new creative strategies. Longer-term studies could also help distinguish whether the system genuinely supports intent development or merely allows participants to gradually learn how to adapt to the model.

From a methodological perspective, this paper uses Flow-GRPO to directly connect artists' trajectory-level judgments to generative model updates, providing one concrete implementation for sustained personalization, but the contribution of this paper does not depend on any specific reinforcement learning algorithm. Different generative model architectures, online optimization methods, or parameter-efficient update mechanisms may differ in training stability, response speed, and interaction cost. Future work can further investigate which model update mechanisms are most suitable for interactive personalization, especially because in real artistic creation, users are unlikely to accept long waiting times for model training to finish. Therefore, reducing the computational cost of each round of adaptation and achieving near-real-time personalized updates while maintaining generation quality are important technical challenges for bringing this type of system into real creative practice.

Finally, this paper understands artistic intent as a dynamic object that can gradually emerge and change through interaction, but this view itself also means that the system should not attempt to `infer'' a single and stable true intent. What the model observes is always only the choices expressed by the artist at a particular time and within a particular creative context. Future systems need to avoid presenting probabilistic preference patterns as deterministic understandings of a creator's identity. Rather than claiming that the model already `knows who the artist is,'' a more reasonable goal may be to build a continuously revisable collaborative relationship in which the system provides more relevant possibilities based on existing selections while always allowing the artist to change direction.

Overall, this paper demonstrates a sustained co-creation process from exploration to recognition, then to shaping and re-exploration, but this loop is still only a starting point for understanding long-term personalized co-creation. Future research needs to further explore how generative systems can remember creators without fixing them in the past, and how they can learn individual preferences while continuously preserving space for new creative intentions. For generative AI designed for artistic practice, achieving this balance may be more important than simply pursuing higher preference alignment.

\section{Conclusion}

Generative AI is gradually shifting from a content generation tool that executes explicit instructions to a creative medium that can continuously participate in creators’ processes of thinking, judgment, and exploration. Building on this shift, this paper focuses on a question that remains insufficiently discussed in existing research on generative interaction: when artists’ creative intentions are not fully formed before interaction begins, and may still remain latent, vague, or difficult to articulate, how should generative systems support the gradual discovery of such intentions and further allow them to influence the development of the model itself?

To address this question, we propose a sustained human--AI co-creation framework composed of
\textit{exploration--recognition--shaping--re-exploration}.
Rather than understanding exploration solely as an internal model search mechanism, we reinterpret it as a creator-oriented
\textit{intent discovery mechanism}:
by actively expanding the visible generation space, the model enables artists to encounter visual possibilities beyond the scope of their current explicit expressions, and to gradually identify creative directions that resonate with them through observation, comparison, and curation. We further connect these curatorial judgments directly to the reinforcement learning process, translating artists’ degrees of preference for different generation trajectories into different learning strengths. In this way, preference discovery does not remain limited to the current generation round, but can continuously shape the model’s future generation behavior.

This mechanism also changes the meaning of personalization in generative creation. This paper does not assume that artists must first define a stable and complete preference profile for the model to fit. Instead, individual preferences can gradually emerge during actual creative practice, while the model can continuously adapt to these emerging judgments within the same process. Therefore, what develops between the artist and the model is not a one-way process of ``user feedback $\rightarrow$ model optimization,'' but a sustained process of co-adaptation: the model helps the artist encounter possibilities that were not previously explicit, while the artist’s selections further determine where the model is more likely to continue exploring next.

Around this framework, we conduct three user studies to examine whether exploration can help artists discover latent preferences, whether artist curation can improve artist-specific preference alignment, and whether sustained human--AI co-creation can further help creators externalize and materialize creative intentions that are latent, vague, and difficult to clearly articulate. By situating preference discovery, model personalization, and long-term co-creation within the same research framework, this paper seeks to extend the focus of generative interaction from `how accurately can a system execute an articulated intent'' to `how can a system participate in the formation and development of intent itself.''

More broadly, we argue that generative systems designed for creative practice should not merely pursue increasingly accurate predictions of what users have liked in the past. A human--AI co-creation system that genuinely supports creativity also needs to preserve space for preferences that have not yet formed, while allowing users to reinterpret, revise, or even change their creative directions through sustained interaction. Therefore, a key question for future personalized generative systems may not be how quickly they can converge toward a stable user model, but how they can maintain a dynamic balance between remembering creators and continuing to present them with new possibilities. The framework proposed in this paper provides a starting point for this direction: artistic intent is no longer only an input that must be fully specified before generation begins, but can become an object that is gradually discovered, externalized, and further shaped through sustained co-creation between people and models.

\clearpage
\bibliographystyle{ACM-Reference-Format}

\bibliography{refer.bib}

% \clearpage

% \end{CJK*}
\end{document}